\documentclass{article}

\usepackage[verbose=true,letterpaper]{geometry}
\usepackage{pgfplots}
\usepackage{subfig}
\usepackage{arxiv}

\usepackage[utf8]{inputenc} 
\usepackage[T1]{fontenc}    
\usepackage{hyperref}       
\usepackage{url}            
\usepackage{booktabs}       
\usepackage{amsfonts}       
\usepackage{nicefrac}       
\usepackage{microtype}      
\usepackage{lipsum}
\usepackage{graphicx}
\usepackage{amsfonts}
\usepackage{array}
\newcolumntype{C}[1]{>{\centering\let\newline\\\arraybackslash\hspace{0pt}}m{#1}}

\title{How to read faces without looking at them}

\author{
  Suyash Shandilya \\
  Defence Research and Development Organisation\\
  New Delhi - 110054\\
  \texttt{su.sh2396@gmail.com} \\
   \And
 Waris Quamer \\
  Amdocs\\
  Pune, Maharashtra - 411028\\
  \texttt{mr.warisquamer@gmail.com} \\
}

\begin{document}
\maketitle

\begin{abstract}
Face reading is the most intuitive aspect of emotion recognition. Unfortunately, digital analysis of facial expression requires digitally recording personal faces. As emotional analysis is particularly required in more poised scenario, capturing faces becomes a gross violation of privacy. In this paper, we use the concept of \emph{compressive analysis} introduced in \cite{mima} to conceptualise a system which compressively acquires faces in order to ascertain unusable reconstruction, while allowing for acceptable (and adjustable) accuracy in inference.
\end{abstract}

\keywords{Compressive analysis \and Privacy \and Emotion detection \and Compressed sensing \and Machine Learning}

\section{Introduction}\label{sec:intro}

\paragraph{Need for EQ}
As we move towards advancing artificial intelligence, we need to realise that intelligence is more than logical intellect. It is the ability to accurately perceive a data in a given context, and return an apposite response. Faster machines will render faster results, smarter machines will render smarter result. But the \textit{hoi polloi} doesn't always seek an optical solution to a complex computational problem. Intuition entices more than intelligence. An 'IQ' alone cannot suffice to build human-like intuition. \textbf{Emotional Quotient} (EQ) is a major pillar for the foundation of future technology. Advances in deep learning have already shown good promise in the domain; what is required is an intent to accomplish it.

\paragraph{Need for Privacy}
Reading faces is one of the most trivial ways humans perceive each other's emotions. Our ability to express and read emotions via face and other physical cues is what advanced our species in communication. Other physiological signals like pulse rate, breathing pattern, also convey a more objective emotional analysis. EEG signals would probably top all of them but none of these can be as conveniently acquired as capturing facial expression. Nevertheless, all of these are privacy invasive. There is no prima facie way to infer something as personal as emotion, without invading some amount of privacy. The concept of \textit{compressive analysis} allows one to seek a trade-of between privacy and personalisation.

\subsection{Compressive analysis}\label{sec:ca}
Compressive Sensing \cite{donoho,candromtao} is a recent idea which includes compression as a part of acquisition itself by acquiring random linear measurements instead of uniform samples. It further defines a lower bound on the number of compressed samples one needs to acquire to reconstruct the original data accurately; or to an accuracy limited only by the level of noise in acquisition. Interested readers are recommended to read the excellent introduction on the topic given by Candes and Wakin \cite{cstut}. As the number of acquired measurements are reduced further, the reconstruction quality degrades to a level of unusability very soon. Even from such an irretrievable state, one can expect - given the art of compressive sampling - certain key structures to be preserved deep in the randomness of the compression. The idea of compressive analysis is to analyse this compressed representation (similar to a non-cryptographic digest) of the original image instead of the image itself. For a sufficient degree of compression, a reconstruction may not render a sensible image but still be utilisable for certain important analysis (emotion detection, in this work).
It maybe noted that since the analysis is performed only on the compressed version of the image, any reconstruction requires the knowledge of the sensing matrix $\phi$ which is (in our case) a binary gaussian random matrix of size $M \times N$ where $M$ are the number of compressed samples and $N$ is the length of the image vector (concatenated image matrix). Thus even if the reconstruction may render a 'sensitive' information in some sense, it can only be recovered by someone who has the knowledge of the matrix $\phi$.
\subsection{Previous work} \label{sec:prev}
This is a more advanced demonstration of the idea of compressive analysis first broached in \cite{mima}. As stated there, the concept of compressive classification was posited along with the single-pixel camera \cite{spc} by the name of smashed filters \cite{smash}. \cite{ppfacerec} demonstrates privacy preserving face recognition using secure multiparty computation. The idea in the paper was to have privacy preserving biometric verification system. Homomorphic encryption is a common idea in \cite{ppfacerec,ppfacerec2} and many similar approaches to the idea. To the best of our research, this is the only paper so far, that assures privacy preserving emotion recognition from pure image analysis.

\subsection{Paper Layout} \label{sec:layout}
The rest of the paper is covered as follows. Section \ref{sec:method} details the specifics of the methodology employed including the chosen dataset followed by the preprocessing (encryption/compression in our case) specifications. Section \ref{sec:results} presents the results of the chosen classification models and analyses their performances. Section \ref{sec:reconst} describes the reconstruction from the compressed images. Section \ref{sec:conclusion} puts forward the conclusion and lays down the future scope of our research.

\section{Methodology}\label{sec:method}
\subsection{Dataset}\label{sec:data}
\begin{figure}[ht]
    \centering
    \includegraphics[scale = 0.5]{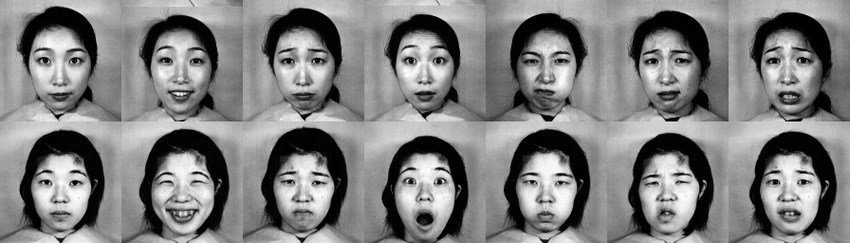}
    \caption{Samples from the Japanese Female Facial Expression (JAFFE) dataset}
    \label{fig:imfdb}
\end{figure}
The Japanese Female Facial Expression (JAFFE) database contains 213 images of 7 facial expressions (6 basic facial expressions + 1 neutral) posed by 10 Japanese female models. Each image has been rated on 6 emotion adjectives by 60 Japanese subjects.
\subsection{Image Encryption}\label{sec:enc}
Compressive Sensing requires that the vector being recovered should be sparse in some domain. The Daubechies-10 \cite{db} wavelets were chosen for a sparse representation of the images here. The images underwent a 2D wavelet transformation before being concatenated to a column vector. As earlier in \cite{mima}, binary gaussian matrix was chosen as a sensing matrix $\phi \in \{0,1\}^{M \times N}$ to simulate the DMD in the single-pixel camera \cite{spc}. For $M = 2000$, \emph{some} reconstructions showed some characterisable facial features, thus it was decided that M must be less than 2000. For respectable classification accuracy, the lower bound for $M$ was chosen as 50. To observe of the behaviour of the model for even lower values of M, the final values were chosen as: 800,500,200,100,50,20,10,5,2,1. $M = 50$ is equivalent to a compression ratio of $0.31\%$ ($800 \equiv 5\%$).

\subsection{Experimental Procedure} \label{sec:exp}
\begin{figure}[ht]
    \centering
    \includegraphics[scale = 0.8]{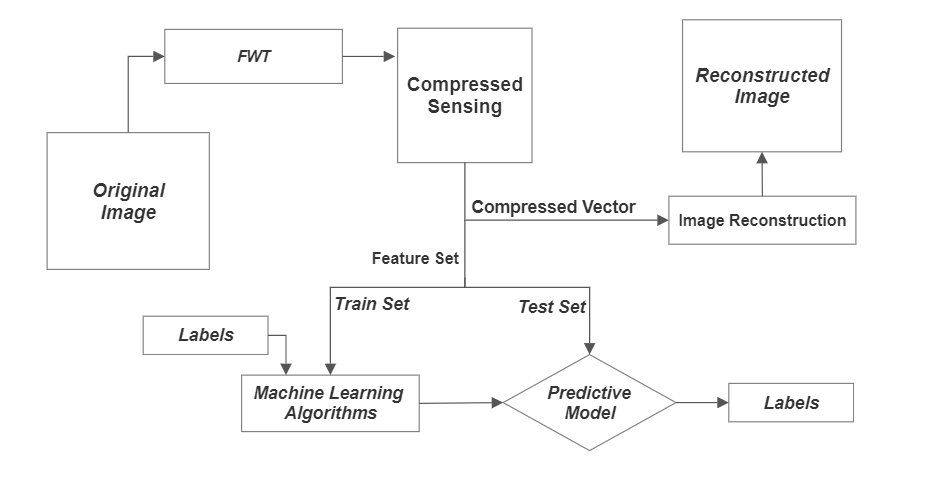}
    \caption{Systematic representation of our proposed work}
    \label{fig:flow}
\end{figure}

Each image from the dataset was read as 128x128 matrix. Fast Wavelet Transform (FWT) was then applied on the images obtaining the discrete wavelet coefficients at the maximum decomposition level (5 in this case) for the \textit{db2} wavelet filter-bank. The coefficients were stacked to form the image vector of size \textit{$N \times 1$}, which was multiplied with the sensing matrix $\phi$ of size \textit{$M \times N$} (as mentioned in section \ref{sec:ca} and \ref{sec:enc}) to obtain the compressed vector of the image. This compressed vector was used as the feature set for the classification models as well as the image reconstruction (see section \ref{sec:reconst}). The systematic representation of our proposed work is explained in Figure \ref{fig:flow}.

\section{Results and Analysis}\label{sec:results}
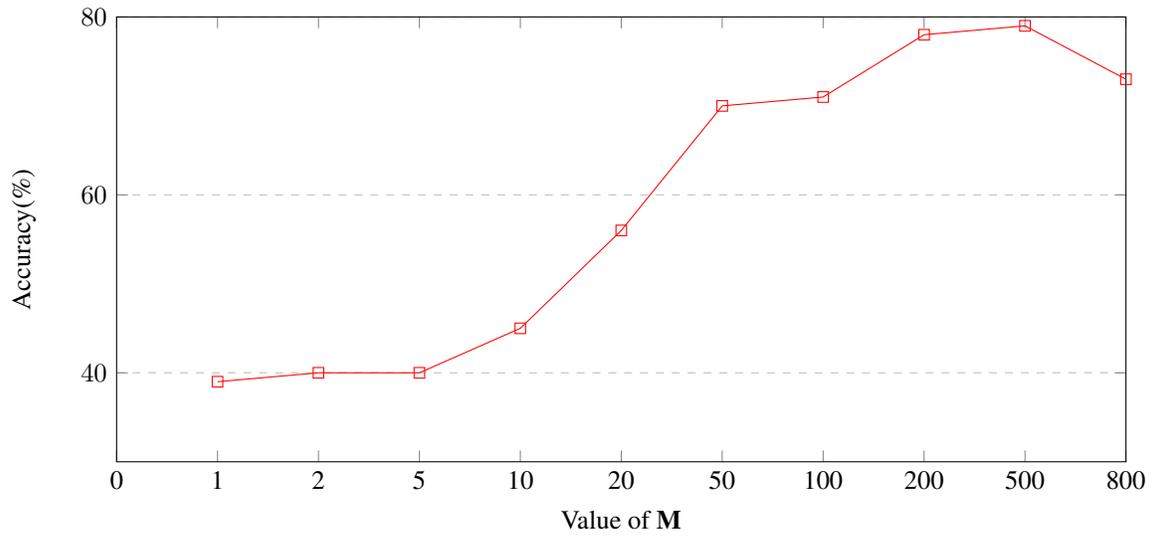
\begin{figure}
\begin{center}
\begin{tikzpicture}
\begin{axis}[
    width = 15cm,
    height = 7.5cm,
    xlabel={Value of \textbf{M}},
    ylabel={Accuracy(\%)},
    xmin=0, xmax=10,
    ymin=30, ymax=80,
    xtick={0,1,2,3,4,5,6,7,8,9,10},
    xticklabels ={0,1,2,5,10,20,50,100,200,500,800},
    ytick={20,40,60,80},
    ymajorgrids=true,
    grid style=dashed,
]

\addplot[
    color=red,
    mark=square,
    ]
    coordinates {
    (1,39)(2,40)(3,40)(4,45)(5,56)(6,70)(7,71)(8,78)(9,79)(10,73)
    };

\end{axis}
\end{tikzpicture}
\end{center}
\caption{Accuracy of MLP Model Vs different Values of M}
\label{fig:accuracy}
\end{figure}

 For our analysis, the JAFFE dataset was classified into three broad emotional categories, as Positive, Neutral and Negative. Different classification models namely, Multilayer Perceptrons(MLPs), k-nearest neighbor, Decision tree, Support vector machines among others, were tested with various different parameter values. For evaluating the performance of the classifiers, a 5-fold cross validation accuracy score was used for the value of M = 500. The performances of different models for their optimised parameter values are shown in Table \ref{table:all}. With the ability to dynamically model non-linear and complex prediction functions, MLPs can learn hidden relationships without imposing any restrictions on the data. It is evident from the results that Multilayer Perceptron (MLP) was the best performing model whereas the rest of the classifiers had notably low accuracies as compared to that of the neural nets. Therefore, it was concluded that the MLPs have a better learning capacity and thus better suited for this problem. The performance of MLP was then further analysed for different values of M, the results of which are depicted in Table \ref{table:mlp} and Figure \ref{fig:accuracy}. It can be observed that the accuracy of the model increases when the value of M is reduced from 800 to 500. One of the probable reasons for this behaviour could be a high correlation between attributes values for M = 800. For the lower values of \textbf{M} (i.e $M  \in \{1,2,5\}$), the model seemed to be biased towards a particular class. There must have been too much loss in data for the model to be able to differentiate between classes.

\begin{table}[ht]
\centering
\begin{tabular}{|c|c|}
\hline
\textbf{Model}           & \textbf{Accuracy} \\ \hline
K-Nearest Neighbors      & 0.49              \\ \hline
SVM (Linear kernal)      & 0.42              \\ \hline
SVM (RBF kernal)         & 0.42              \\ \hline
Gaussian Process         & 0.44              \\ \hline
Decision Tree            & 0.28              \\ \hline
Random Forest            & 0.26              \\ \hline
\textbf{Multilayer Perceptron} & \textbf{0.79}     \\ \hline
AdaBoost                 & 0.44              \\ \hline
Naive Bayes              & 0.30              \\ \hline
QDA                      & 0.30              \\ \hline
\end{tabular}
\caption{5-Fold Cross Validation Score for Classifiers for \textbf{M = 500} }
\label{table:all}
\end{table}

\begin{table}[ht]
\centering
\begin{tabular}{|C{1.5cm}|c|c|c|}
\hline
\textbf{M}  & \textbf{Compression Ratio(\%)} & \textbf{Train Accuracy} & \textbf{Test Accuracy} \\ \hline
800     & 4.89      &0.90   & 0.73     \\ \hline
\textbf{500}     & \textbf{3.05}      &\textbf{0.97} & \textbf{0.79}     \\ \hline
200     & 1.22      &0.96   & 0.78     \\ \hline
100     & 0.61      &0.95   & 0.71     \\ \hline
50      & 0.31      &0.91   & 0.70     \\ \hline
20      & 0.12      &0.74   & 0.56     \\ \hline
10      & 0.06      &0.62   & 0.50     \\ \hline
5       & 0.03      &0.52   & 0.40     \\ \hline
2       & 0.01      &0.43   & 0.40     \\ \hline
1       & 0.006     &0.43   & 0.39     \\ \hline
\end{tabular}
\caption{Accuracy of MultiLayer Perceptron for Different Values of \textbf{M} }
\label{table:mlp}
\end{table}

\section{Image Reconstruction} \label{sec:reconst}
\begin{figure}
\centering
\begin{minipage}{.24\textwidth}
\centering
\subfloat{{\includegraphics[width=2.5cm]{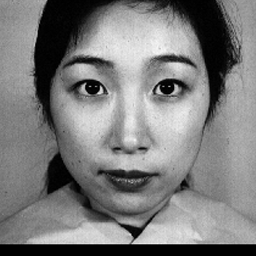} }}%
\qquad
\subfloat{{\includegraphics[width=2.5cm]{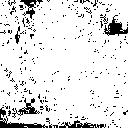} }}%
\qquad
\subfloat{{\includegraphics[width=2.5cm]{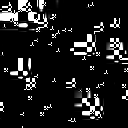} }}%
\qquad
\subfloat{{\includegraphics[width=2.5cm]{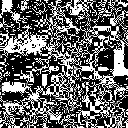} }}%
\end{minipage}%
\begin{minipage}{.24\textwidth}
\centering
\subfloat{{\includegraphics[width=2.5cm]{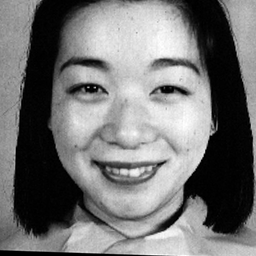} }}%
\qquad
\subfloat{{\includegraphics[width=2.5cm]{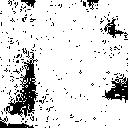} }}%
\qquad
\subfloat{{\includegraphics[width=2.5cm]{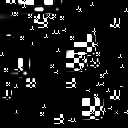} }}%
\qquad
\subfloat{{\includegraphics[width=2.5cm]{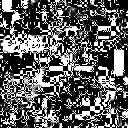} }}%
\end{minipage}
\begin{minipage}{.24\textwidth}
\centering
\subfloat{{\includegraphics[width=2.5cm]{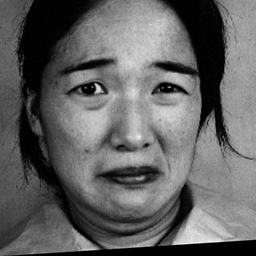} }}%
\qquad
\subfloat{{\includegraphics[width=2.5cm]{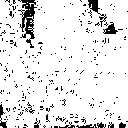} }}%
\qquad
\subfloat{{\includegraphics[width=2.5cm]{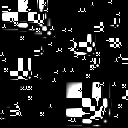} }}%
\qquad
\subfloat{{\includegraphics[width=2.5cm]{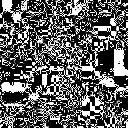} }}%
\end{minipage}
\begin{minipage}{.24\textwidth}
\centering
\subfloat{{\includegraphics[width=2.5cm]{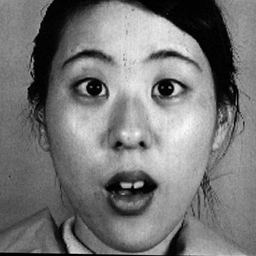} }}%
\qquad
\subfloat{{\includegraphics[width=2.5cm]{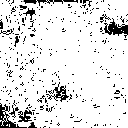} }}%
\qquad
\subfloat{{\includegraphics[width=2.5cm]{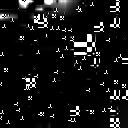} }}%
\qquad
\subfloat{{\includegraphics[width=2.5cm]{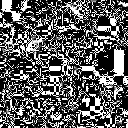} }}%
\end{minipage}
\caption{The first row consists of samples from our dataset for different labels (L to R - neutral, positive, negative, negative) and susequent rows are the reconstruction results for $M = 500,50,10$ respectively.}
\label{fig:reconst}
\end{figure}

Typically, Matching Pursuit algorithms \cite{mps} are more popular in use because of the speed and ease in execution. Given our objective here is to ensure maximum security of the target data, we have stuck to using basis pursuit \cite{bpdn} for all our comparative reconstruction because of stronger guarantees of accuracy, albeit at the cost of computation capital. The SPGL1 \cite{spgl1} library was used to reconstruct the images. Since noise will only obfuscate reconstruction accuracy, and a well trained classifier can typically work as good in its presence, we did not consider any acquisition noise in our experiments. It was also noted that CVX \cite{cvx,cvx2} gave slightly better reconstruction but was about 4 times slower.
The images in Fig \ref{fig:reconst} may seem as a glitch to a passing reader but they are the best possible reconstruction we have been able to present. No single colormap could produce a more comprehensible image. One of the other shortcomings of Compressed sensing in general is that it will require  very high bit depth to better portray the computation results. The checkered artefacts (more prominent in the bottom 2 rows) evince a very sparse vector in the wavelet basis.
One may argue that better reconstruction is possible if we regularise the objective function with the $l_2$-norm. That may be true but it seemed practically impossible for us in our experiments. Moreover, the regularisation manifests more as noise than as clarity (if any).
Nevertheless, there is a considerable room for improvement which seems more befitting to be mentioned in the next section.
\section{Conclusion and Future Works} \label{sec:conclusion}
We have successfully demonstrated how - using a well trained classifier - extremely compressed samples which are unusable for a recognisable reconstruction, can be used to for classification even in subtle privacy sensitive applications like recognising facial expressions. Although only MLPs performed up to the mark in our analysis, other simple models can be reasonably effective if the degree of compression is reduced.
The dataset used in our analysis has been made out of good and earnest efforts. The consistency of images makes computation easy. Unfortunately the real life scenario may be far from it. Our aim here was to prove the concept for face portraits. In future work, we will attempt to develop systems which are able to be at least as efficient on a more diverse and larger dataset.
Significant improvement in reconstruction can be achieved by changing the basis of image representation to other more compatible bases. Intuitively it can be understood, how Ridgelets, or Curvelets \cite{curve} will be more compliant with portrait images than wavelets per se. More sophisticated splines can be developed to achieve an even better transformation. The intuition is to capture more structure of the image in the individual basis vectors. The same idea can be extended to the optimisation problem by utilising the block sparsity of the image vector in the wavelet basis.
Visualising way ahead in time, if such a unit is deployed on a large scale, it could lead to having multiple different measurements of a person with the same expression. An unwanted adversary, who somehow manages to assimilate all these different measurements along with the sensing matrices may attempt to reconstruct certain faces. Even if a reconstruction seems very overreaching, a detection from a known subset of people could be feasible. In future we wish to analyse whether, and to what extent, such classifications are possible from compressed samples.
\bibliographystyle{unsrt}  
\bibliography{references}

\end{document}